# Surface Light Field Fusion

Jeong Joon Park
University of Washington
jjpark7@cs.washington.edu

Richard Newcombe
Facebook Reality Labs
newcombe@fb.com

Steve M. Seitz
University of Washington
seitz@cs.washington.edu

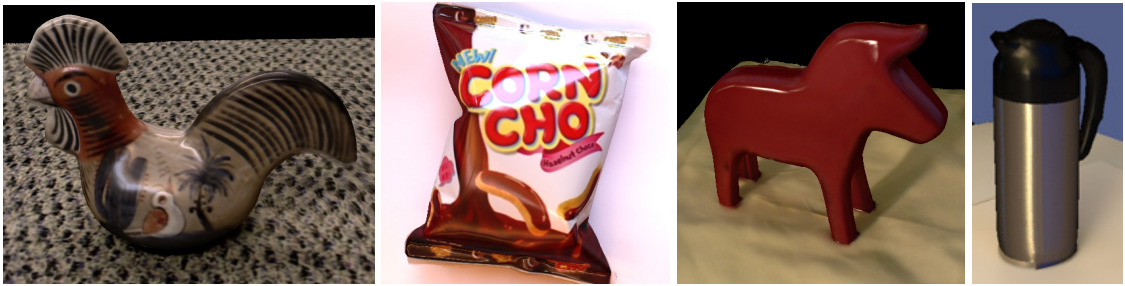

Figure 1: Synthetic renderings of highly reflective objects reconstructed with a hand-held commodity RGBD sensor. Note the faithful texture, specular highlights, and global effects such as interreflections (2nd image) and shadows (3rd image).

## Abstract

*We present an approach for interactively scanning highly reflective objects with a commodity RGBD sensor. In addition to shape, our approach models the* surface light field, *encoding scene appearance from all directions. By factoring the surface light field into view-independent and wavelength-independent components, we arrive at a representation that can be robustly estimated with IR-equipped commodity depth sensors, and achieves high quality results.*

## 1. Introduction

The advent of commodity RGBD sensing has led to great progress in 3D shape reconstruction [9, 30, 40]. However, the *appearance* of scanned models is often unconvincing, as view-dependent reflections (BRDFs) and global light transport (shadows, interreflections, subsurface scattering) are very difficult to model.

Rather than attempt to fit BRDFs and invert global light transport, an alternative is just to model the radiance (in every direction) coming from each surface point. This representation, known as a *surface light field* [41], captures the scene as it appears from any viewpoint in its native lighting environment.

We present the first Kinect Fusion-style approach designed to recover surface light fields. As such, we are able to convincingly reproduce view-dependent effects such as specular highlights, while retaining global effects such as diffuse interreflections and shadows that affect object appearance in its native environment (Fig. 1). To achieve this capability, we approximate the full surface light field by factoring it into two components: 1) a local specular BRDF model that accounts for view-dependent effects, and 2) a diffuse component that factors in global light transport.

The resulting approach reconstructs both shape and high quality appearance, with about the same effort needed to recover shape alone – we simply require a handful of IR images taken during the usual scanning process. In particular, we exploit the IR sensor in commodity depth cameras to capture not just the shape, but also the view-dependent properties of the BRDF in the IR channel. Our renderings under environment lighting capture glossy and shiny objects, and provide limited support for anisotropic BRDFs.

We make the following contributions. The first is our novel, factored, surface light field-based problem formulation. Unlike traditional BRDF-fitting methods which enable scene relighting, we trade-off relighting for the ability to capture diffuse global illumination effects like shadows and diffuse interreflections that significantly improve realism. Second, we present the first end-to-end system for capturing full (non-planar) surface light field object models with a hand-held scanner. Third, we introduce the first reconstructions of *anisotropic* surfaces using commodity RGBD sensors. Fourth, we introduce a high-resolution texture tracking and modeling approach that better models high frequency details, relative to prior Kinect Fusion [30] approaches, while retaining real-time performance. Fifth, we apply semantic segmentation techniques for appearance modeling, a first in the appearance modeling literature.

## 2. Problem Formulation and Related Work

The input to our system is an RGBD stream and infrared (IR) video from a consumer-grade sensor. The output is a shape reconstruction and a model of scene appearance represented by a surface light field $SL$, which describes the radiance of each surface point $x$ as a function of local outgoing direction $\omega_o$ and wavelength $\lambda$: $SL(x, \omega_o, \lambda)$. We introduce a novel formulation designed to effectively represent and estimate $SL$. We start by writing $SL$ in terms of the rendering equation [17]:

$$SL(x, \omega_o, \lambda) = \int_\Omega E(x, \omega_i, \lambda) f(x, \omega_o, \omega_i, \lambda)(n_x \cdot \omega_i) d\omega_i, \tag{1}$$

where $E$ denotes global incident illumination, $f$ surface BRDF, $\omega_i$ incident direction, and $n_x$ surface normal.

Prior work on modeling $SL$ uses either parametric BRDF estimation methods or nonparametric image-based rendering methods.

BRDF estimation methods [12, 22, 42, 43] solve for an analytical BRDF $f$ by separating out the global lighting $E$. Accurately modeling the global light transport, however, is extremely challenging; hence most prior art [12, 22, 42, 43] simply ignore global effects such as shadows and interreflections, assuming the scene surface is convex. Occlusions and interreflections cause artifacts under this assumption, so some methods [42, 43] require a black sheet of paper below the target object to minimize these effects.

Image-based rendering techniques [5, 28, 41] avoid this problem by nonparametrically modeling the surface light field $SL$, via densely captured views on a hemisphere. However, the dense hemispherical capture requirement is laborious, and therefore not suited to casual hand-held scanning scenarios.

We address these two issues by 1) *empirically* capturing the diffuse component of the appearance that factors in the global lighting, and 2) *analytically* solving for *specular* parameters leveraging the built-in IR projector.

Specifically, we further decompose $SL$ into a diffuse term $D$, specular term $S$ with *wavelength-independent* BRDF (as in the Dichromatic model [33]), and residual $R$:

$$SL(x, \omega_o, \lambda) = \underbrace{\int_\Omega E(x, \omega_i, \lambda) f_d(x, \lambda)(n_x \cdot \omega_i) d\omega_i}_{D(x, \lambda)}$$
$$+ \underbrace{\int_\Omega E_d(x, \omega_i, \lambda) f_s(x, \omega_o, \omega_i; \beta)(n_x \cdot \omega_i) d\omega_i}_{S(x, \omega_o, \lambda)}$$
$$+ R(x, \omega_o, \lambda), \tag{2}$$

where $f_d$ is diffuse albedo, $f_s$ specular component of BRDF with parameters $\beta$, and $E_d$ direct illumination; we ignore specular interreflection.

Rather than infer diffuse albedo $f_d$ as in [12, 22, 42, 43], we simply capture the diffuse appearance $D$ directly in a texture map (the concept referred to as a "lightmap" in the gaming industry). We thereby make use of the rich global effects like shadows and interreflections that are captured in $D$ but avoid the difficult problem of factoring out reflectance and multi-bounce lighting. In doing so, we give up the ability to relight the scene, in exchange for the ability to realistically capture global illumination effects.

The specular term $S$ is approximated by recovering a parametric BRDF $f_s$ with an active-light IR system, assuming that $f_s$ is wavelength-independent. $S$ captures the specular properties of *dieletric* materials like plastic or wood and non-colored metals like steel or aluminum [33, 36, 46].

We assume the residual $R$ to be negligible - as such, our approach does not accurately model colored metals like bronze or gold, and omits specular interreflections.

Overall, our formulation has two main advantages. First, we realistically capture diffuse global illumination effects in $SL$ without explicitly simulating global light transport. Second, the *specular* BRDF is analytically recovered from only a sparse set of IR images, removing the need for controlled lighting rigs [14, 22, 46] and extensive view sampling [5, 15, 28, 41].

***Other Related Works*** Some authors explicitly model global light transport, as in [24] and, to a lesser extent, [32] that assumes low-frequency lighting and material. However, this requires solving a much more difficult problem, and the authors note difficulties in recovering highly specular surfaces and limit their results to simple, mostly uniform textured objects.

[15] captures a *planar* surface light field with a monocular camera using feature-based pose tracking. Although [15] removes the need for special gantries used in [23, 41], it still requires a dense sampling of the surface light field, making it hard to scale beyond a small planar surface.

A few other authors utilize IR light and sensors to enhance the estimation of surface properties. [6] observes that many interesting materials exhibit limited texture variations in the IR channel to estimate the geometric details and BRDF. [7] refines a reconstructed mesh with shading cues in the IR channel. [18] leverages an IR sensor to add constraints to the intrinsic image decomposition task.

## 3. System Overview

Our system operates similarly to Kinect Fusion [30] in that a handheld commodity RGBD sensor is moved over a scene to reconstruct its geometry and appearance. The diffuse component $D$ is obtained by interactively fusing the high resolution texture onto the base geometry [30] and

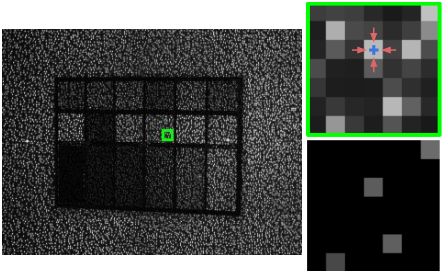

Figure 2: IR image (left), zoomed in (top-right) and local maximum subsampled image (bottom-right). Best viewed digitally.

subsequently removing baked-in specular highlight using a min-composite [35] approach. (Sec. 5)

Section 4 describes recovering $S$. We use a learning-based classifier to segment the scene with similar surface properties and estimate the specular BRDF $f_s$ for each segment using a calibrated IR system. $S$ is then computed by taking the product of the recovered $f_s$ and environment lighting captured from a low-end $360°$ camera.

## 4. Computing $S$

In this section, we assume that the geometry is already reconstructed and focus on estimating $S$, the view-dependent component of the surface light field. We assume that the material varies over the surface of the object or scene, but that there is a relatively small number of material clusters which estimates can be aggregated. We begin by calibrating the infrared (IR) system attached to the depth sensor and describe how to measure material properties in the IR channel for each material.

### 4.1. IR Projector Modeling

***BRDF Estimation in the IR Channel*** We leverage the IR projector built-in to most depth sensors as a point light source. Because we know the projector location relative to the camera along with the scanned geometry, the surface BRDF can be estimated.

One challenge, however, is that the IR projector is typically not uniform – we use a Primesense structured light sensor that generates complex speckle patterns (Fig. 2).

We take a simple approach to select pixels that are lit by the projector: we divide the IR image into a grid structure and keep the brightest pixel within each 5x5 grid (Fig. 2). The intensity of such pixel is averaged with its four-neighborhood. Throughout the paper, a *pixel* in the IR image refers to this local maximum intensity. For simplicity, we consider only the central 192x192 region, to minimize the effects of vignetting and radial distortion.

Despite our usage of the structured light sensor, our system can be easily generalized to other depth sensors with a projector and a receiver pair. For example, we refer readers to [18] for a time-of-flight projector model.

***IR Projector and Camera Calibration*** With constant exposure time, the IR projector's light intensity $\kappa$ and the camera's gamma compression parameter $\gamma$ is optimized as following by capturing a white piece of paper whose albedo is assumed to be 1:

$$\min_{\kappa,\gamma} \sum_{\boldsymbol{x}\in P} \left( L(\boldsymbol{x}) - \left(\kappa \frac{\boldsymbol{n_x} \cdot \boldsymbol{l_x}}{\pi d_{\boldsymbol{x}}^2}\right)^{\gamma} \right)^2, \quad (3)$$

where $L(\boldsymbol{x})$ is observed IR intensity at a pixel $\boldsymbol{x}$ in a collated pixel set $P$ over multiple images, and $\boldsymbol{n_x}, \boldsymbol{l_x}$ and $d_{\boldsymbol{x}}$ are normal, light direction and distance to the light source of the surface point seen by $\boldsymbol{x}$, respectively. Following [7], we assume indoor ambient IR light is negligible.

### 4.2. Per-Segment BRDF Estimation

***Specular Reflection Model*** Renaming the directions $\boldsymbol{\omega_o}$ and $\boldsymbol{\omega_i}$ in Eq. (2) as local view $\boldsymbol{v_x}$ and light directions $\boldsymbol{l_x}$ respectively, we denote the specular component of the BRDF as $f_s(\boldsymbol{v_x}, \boldsymbol{l_x}; \boldsymbol{\beta})$, where $\boldsymbol{\beta}$ is a vector of BRDF parameters that decides the reflectance distribution. In this work, we use the Ward BRDF model [39] for its simplicity and applicability to a variety of materials.

***Material Segmentation*** Accurately fitting BRDF parameters for each individual point requires observing specular highlights for all surface points, which is prohibitively expensive. Therefore, estimating a single BRDF for a region with shared reflectance properties is essential for a practical and robust system. To identify regions of the surface with similar material, we apply an image segmentation method [1] using a Convolutional Neural Network (CNN) for semantic material classification, adapted to operate on a mesh rather than an image (Figure 3). We first convert the patch classifier into a Fully Convolutional Network [25] for dense class predictions (as in [1]) but increase the resolution of the output signal using Dilated Convolution [45]. At each input frame, the output probability map from the softmax layer is projected and averaged into the vertices of our mesh $\mathcal{M} = \{\mathcal{V}, E\}$. We consider the mesh connectivity as a Markov Random Field and obtain the final vertex material label $\boldsymbol{y}$ by minimizing the following energy function:

$$\Phi(\boldsymbol{y}) = \sum_{v\in\mathcal{V}} \psi_v(y_v) + \sum_{\{m,n\}\in E} \psi_{m,n}(y_m, y_n), \quad (4)$$

where the data term $\psi_v(y_v)$ of the MRF is the standard negative log likelihood at vertex $v$: $\psi_v(y_v) = -\log(p_i(y_v))$, and $\psi_{m,n}(y_m, y_n)$ is the pairwise term. Please refer to the supplementary material for description of the pairwise term.

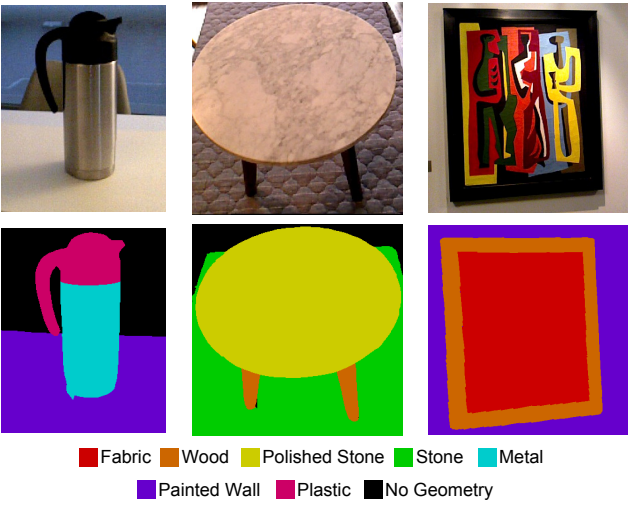

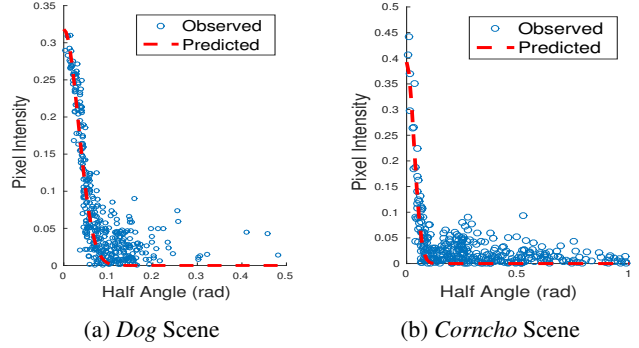

(a) *Dog* Scene  (b) *Corncho* Scene

Figure 4: Specular BRDF fitting results for two scenes, plotting specular components versus half angles of pixel observations along with the fitted prediction curves.

Figure 3: Material segmentation results. The first row shows pictures of target scenes, and the second row shows 3D material segmentation results. Note that the segmentation algorithm works on the reconstructed mesh instead of a single image.

***Specular BRDF Optimization*** For each material segment, the specular parameters $\boldsymbol{\beta}$ that best explain the IR video are estimated. Each pixel $\boldsymbol{x}$ in an IR frame constrains the shared specular parameters $\boldsymbol{\beta}$ and the IR diffuse albedo $\rho(\boldsymbol{x})$ of a surface point. We thus minimize the difference between the actual intensity $L(\boldsymbol{x})$ and the prediction from our model to obtain the optimal $\boldsymbol{\beta}$ and IR diffuse albedo $\boldsymbol{\rho}$:

$$\min_{\boldsymbol{\beta},\boldsymbol{\rho}} \sum_{\boldsymbol{x}\in P} \left( L(\boldsymbol{x}) - \left( \kappa \frac{\boldsymbol{n_x} \cdot \boldsymbol{l_x}}{d_{\boldsymbol{x}}^2} \left( \frac{\rho(\boldsymbol{x})}{\pi} + f_s(\boldsymbol{v}_x, \boldsymbol{l}_x; \boldsymbol{\beta}) \right) \right)^\gamma \right)^2, \quad (5)$$

where the set $P$ collates pixels over all frames. The resulting $\boldsymbol{\beta}$ is used to describe $f_s$, and thus $S$.

The full optimization involves estimating hundreds of thousands of parameters from millions of pixels. So, in practice, we exploit the interactive nature of the scanning process to have the user choose a small subset of frames over which to optimize. These frames are chosen such that a specular highlight is observed in the IR channel and each material is captured at least once. When a material is captured in only one frame, Eq. 5 is solved with spatially constant diffuse albedo. Choosing reference views of convex surface regions improves the specularity estimation by reducing the impact of interreflections. The optimization is solved with Levenberg-Marquardt method [27]. (Fig. 4)

***Anisotropic Surfaces*** Many common reflective surfaces (especially wood and metal) reflect light anisotropically. We therefore extend our approach to model anisotropy for a limited class of surfaces (planes and cylinders) that occur frequently in man-made scenes.

For an anisotropic surface, solving Eq. (5) requires estimating tangent and binormal vectors at each surface point. Among other methods in the literature [11, 38], [14] suggests a photometric approach to estimate the per-point tangent vector by projecting half vectors onto the tangent plane and finding the direction that maximizes symmetry. Applying this technique in our context means densely observing each surface point from many different viewpoints, which is infeasible in a casual hand-held scanning session. We therefore assume that the tangent vector is shared across a material segment, i.e., the surface is planar or cylindrical.

Each pixel $\boldsymbol{x}$ in an IR frame has an associated half-vector $\boldsymbol{h_x}$ and normal vector $\boldsymbol{n_x}$ of the surface point represented in global coordinates and intensity $L(\boldsymbol{x})$. To determine a single representative tangent plane $P_R$, we choose a pixel with the smallest half-angle (angle between half-vector and local normal) as a reference, with its normal $\boldsymbol{n}_R$. Half-vectors of all other pixels relative to its local normal are projected onto the tangent plane of the reference point:

$$P_R(\text{Proj}_{\boldsymbol{n}_R}(\boldsymbol{n}_R + \boldsymbol{h_x} - \boldsymbol{n_x})) \leftarrow L(\boldsymbol{x}). \quad (6)$$

The discretized tangent plane is then smoothed with Gaussian filtering to get the dense BRDF slice as in Figure 5. We then use the Nelder-Mead Simplex Method [29] to find the tangent and binormal vectors that maximize the symmetry [14]. Given the estimated tangent vector, the BRDF parameters $\boldsymbol{\beta}$ for the anisotropic surface are computed through Eq. (5), but with spatially constant diffuse albedo (Fig. 5).

## 5. Computing $D$

In this section, we describe how to obtain high quality scene texture in real-time and iteratively remove baked-in specular highlights as a post-processing step.

**Real-time High Resolution Texture (HRT) Fusion** High quality texture is essential to the perceived realism of scene appearance. While interactive RGBD scanning has seen great progress on delivering quality geometry [9, 40], the state-of-the-art systems still produce low-resolution reconstructions with blurring and ghosting artifacts.

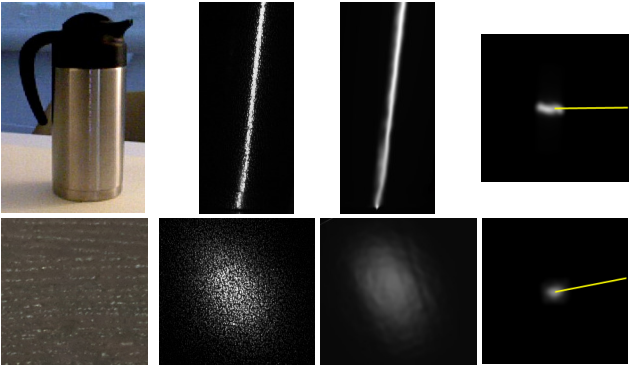

Figure 5: Anisotropic surface fitting results. From left to right: Photo of the target surface (brushed metal cylinder and planar wood surface); Image captured from IR camera; our rendering from the same camera pose; Tangent plane of the reference point where half-vectors of observations are projected (the yellow line is the optimal tangent vector that maximizes symmetry). Best viewed digitally.

Recently, global pose optimization approaches [47] and its patch-based variants [2] have produced impressive texture mapping results. However, these methods work offline and thus cannot provide interactive visual feedback to users on which part of the current model is missing texture or needs close-up scanning.

We show that a high resolution texture representation combined with GPU-accelerated dense photometric pose optimization [19, 40] can greatly improve the real-time scanning quality (Figure 6). We also propose a gradient-based objective function and generalized specular highlight removal algorithm for handling non-Lambertian surfaces.

### 5.1. Preliminaries

We adopt Kinect Fusion [30] to obtain the scene geometry and refine the extracted mesh using the method of [16]. We denote the final mesh, vertex set, and vertex normal set as $\mathcal{M}$, $\mathcal{V}$, and $\mathcal{N}$, respectively.

The focal length $f$ and principal point $c$ of the camera are estimated to form a calibration matrix $K$. We define a projection operator $\pi : \mathbb{R}^4 \mapsto \mathbb{R}^2$ from a 3D homogeneous point to a 2D point on image plane: $\pi(\boldsymbol{p}) := \left( \frac{p_x}{p_z} f_x + c_x, \frac{p_y}{p_z} f_y + c_y \right)$. The rigid body transformation of the camera at time $i$ relative to the first frame is:

$$T_i = \begin{bmatrix} R & t \\ 0 & 1 \end{bmatrix} : \quad R \in \mathbb{SO}3, t \in \mathbb{R}^3, \qquad (7)$$

where a 3D point $\boldsymbol{p}$ in local camera coordinates can be transformed to a point in global coordinates as $\boldsymbol{p}' = T_i \boldsymbol{p}$. We calibrate low polynomial radial and decentering distortion [4] parameters, and each RGB frame is undistorted accordingly. All RGB images are compensated for vignetting after radiometric calibration [20].

### 5.2. Texture Representation

We represent the color texture of the model as a 2D texture atlas [26] $A$ that stores RGB values and associated weights. We parameterize our mesh $\mathcal{M}$ by simply laying each triangle on $A$ without overlap. For each pixel $\boldsymbol{u}$ on the texture plane, its 3D position can be computed through barycentric interpolation, denoted as $\boldsymbol{u}^{3d}$.

### 5.3. Rendering for Appearance Prediction

Using the rasterization pipeline, we render a high resolution prediction of the scene appearance $\mathcal{R}_i$ along with a depth map $Z_i$ and normal map into a camera with estimated pose $T_i$. This high resolution rendering provides constraints for pose optimization for the upcoming input frame.

### 5.4. Texture Update

Given its estimated pose $T_i$, each RGB frame $\mathcal{I}_i$ is densely fused into the texture atlas. We compute the color input of a texture pixel $\boldsymbol{u} \in \Omega$ by projecting the corresponding 3D point $\boldsymbol{u}^{3d}$ onto the bilinearly interpolated image $\mathcal{I}_i$ with depth testing. Specifically, let $\boldsymbol{x}$ be a point on the image plane of $\mathcal{I}_i$ obtained from perspective projection of $\boldsymbol{u}^{3d}$. The color measurement $\mathcal{I}_i(\boldsymbol{x})$ is blended into $A(\boldsymbol{u})$ through weighted averaging with weight $w_i(\boldsymbol{x})$ associated with $\boldsymbol{x}$ at time $i$. We assign high weights to reliable and important color observations:

$$w_i(\boldsymbol{x}) = m_i z_i(\boldsymbol{x}) s_i(\boldsymbol{x}). \qquad (8)$$

Here, $m_i$ accounts for motion blur and rolling shutter effects that occur during fast camera motions measured by the $L^2$ distance between translation and rotation components of $T_{i-1}$ and $T_i$. $z_i(\boldsymbol{x})$ rejects pixels on depth discontinuities where estimated geometry is unreliable. Finally, measurements close to the surface and near perpendicular to it provide sharper texture, yielding the importance term $s_i(\boldsymbol{x})$:

$$s_i(\boldsymbol{x}) = \frac{\boldsymbol{n_x} \cdot \boldsymbol{v_x}}{Z_i^2(\boldsymbol{x})}. \qquad (9)$$

### 5.5. Frame-to-Model Photometric Pose Refinement

To estimate the camera pose $T_i$, we use a frame-to-model RGBD dense alignment approach, minimizing the photometric error between the live frame $\mathcal{I}_i$ and the previous frame's rendering $\mathcal{R}_{i-1}$ on top of traditional dense point-plane Iterative Closest Point algorithm [30].

Unlike previous real-time RGBD tracking methods [19, 40], our 2D texture atlas can store higher frequency photometric details compared to commonly used surfels or volumetric textures, providing tighter pose constraints.

Dense RGBD alignment techniques aim to find a relative rigid body transformation $T^\circ = T_i^{-1} T_{i-1}$ that minimizes photometric error when $\mathcal{I}_i$ is warped into the reference image $\mathcal{R}_{i-1}$ with the corresponding geometry represented by the reference depth map $Z_{i-1}$. For a pixel $\boldsymbol{x}$ on image plane $\Lambda$ of $\mathcal{R}_{i-1}$, we get a 3D point $p_x$ in local coordinates through back-projection: $p_x = Z_{i-1}(x)K^{-1}x$.

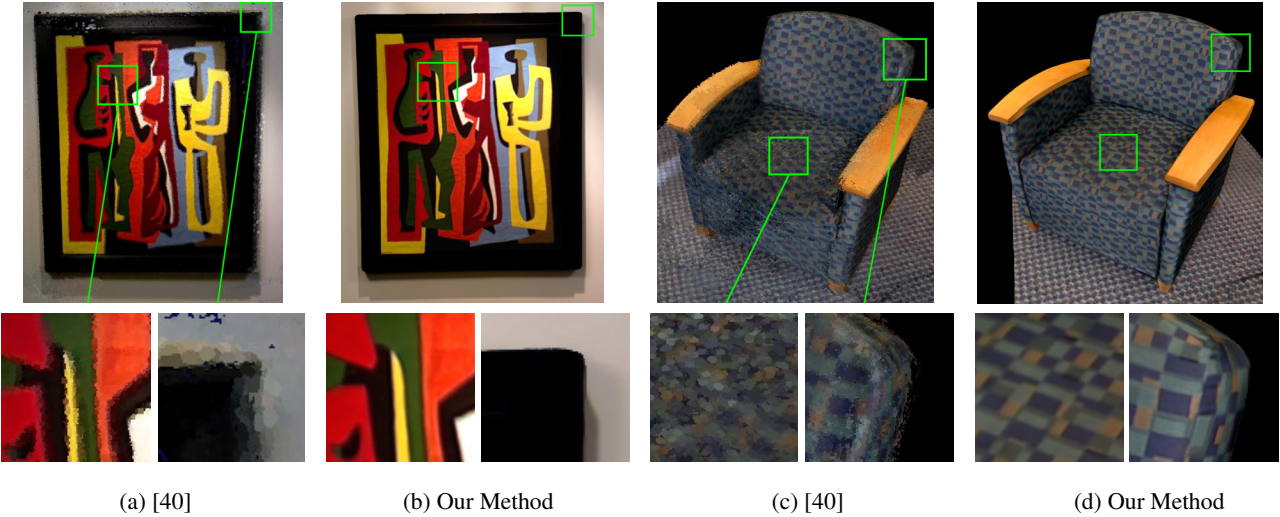

(a) [40]  (b) Our Method  (c) [40]  (d) Our Method

Figure 6: Comparison of texture quality with a state-of-the-art method [40]. The first and third columns show reconstruction results of [40], while the second and fourth show our results. The second row shows magnified patches of the green boxes. Best viewed digitally.

We can then look up how this point appears in the live input image: $\mathcal{I}_i(\pi(T^\circ p_x))$. Modern dense alignment techniques then minimize greyscale photometric error between $\mathcal{R}_{i-1}(x)$ and $\mathcal{I}_i(\pi(T^\circ p_x))$ under a quadratic loss.

For our application, however, maximizing photoconsistency does not necessarily results in accurate camera tracking since specular surfaces can appear significantly different across viewpoints. We find that applying a high-pass filter to images improves tracking accuracy and therefore modify the objective to operate on the gradient of the predicted and live images. The optimal transformation $\hat{T}^\circ$ is then:

$$\hat{T}^\circ = \underset{T^\circ}{\operatorname{argmin}} \sum_{x \in \Lambda} \left( ||\nabla \mathcal{R}_{i-1}^{\mathcal{G}}(x)|| - ||\nabla \mathcal{I}_i^{\mathcal{G}}(\pi(T^\circ p_x))|| \right)^2, \quad (10)$$

where the superscript $\mathcal{G}$ denotes greyscale images.

The intuition is that the high-pass filter reduces each specular highlight (which can be large) to its boundary, and the boundary tends to be lower frequency than the underlying diffuse texture, as the specular BRDF acts as a low-pass filter on the incident illumination [31].

### 5.6. Specular Highlight Removal

The resulting texture has specular highlights "baked-in", which need to be removed to obtain $D$. This problem can be posed as a layer separation problem [8, 44], to separate the stationary diffuse texture from the view-dependent specular layer. Specifically, [35] proposes the *min-composite*, i.e., minimum intensity values across all viewpoints, to approximate (via least upper bound) the diffuse layer.

We use Iteratively Reweighted Least Squares (IRLS) [13] to robustly compute the *min-composite*. For a given

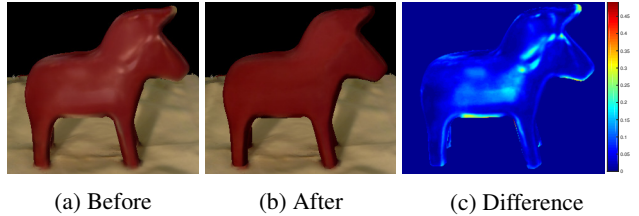

(a) Before  (b) After  (c) Difference

Figure 7: Specular highlights are removed from (a) by computing a *min-composite* via Iteratively Reweighted Least Squares. (b) shows the estimated $D$ and (c) the difference heat map.

pixel on the texture map, computing its weighted average $\hat{q}$ as in Section 5.4 is equivalent to finding the solution of a weighted least squares problem $\operatorname{argmin}_{\hat{q}} \sum_i w_i (q_i - \hat{q})^2$, where $q_i$ is the $i$th intensity input, $w_i$ is the weight of the input computed in equation (8). IRLS filters out the highlights by down-weighting bright outliers in each iteration:

$$\hat{q}^{t+1} = \underset{\hat{q}}{\operatorname{argmin}} \sum_i \mu_{\tau,\upsilon}(\hat{q}^t, q_i) w_i (q_i - \hat{q})^2, \quad (11)$$

$$\mu_{\tau,\upsilon}(\hat{q}^t, q_i) = \begin{cases} 1 & q_i \leq \hat{q}^t + \tau \\ \exp\left(-\frac{(\hat{q}^t + \tau - q_i)^2}{\upsilon^2}\right) & q_i > \hat{q}^t + \tau. \end{cases} \quad (12)$$

In practice, one or two iterations is sufficient. (Figure 7)

While this approach is effective for surfaces with significant diffuse components, it fails for metallic surfaces with negligible diffuse reflection. Fortunately, we can leverage our IR BRDF measurements (Sec.4) to identify such surfaces; we set $D$ to zero for any surface whose estimated diffuse albedo is less than 0.03 and specular albedo greater than 0.15.

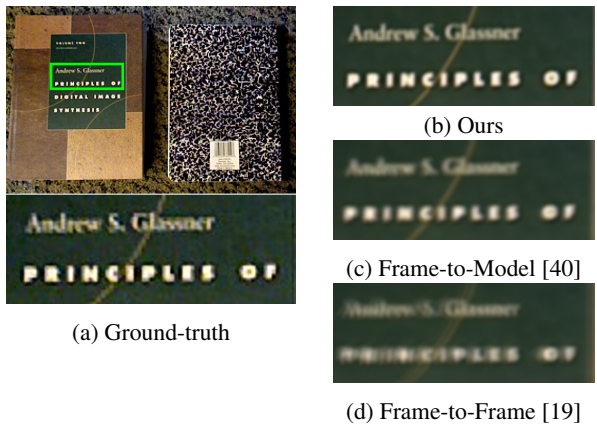

(a) Ground-truth
(b) Ours
(c) Frame-to-Model [40]
(d) Frame-to-Frame [19]

Figure 8: Ground-truth image and predicted rendering of 3D models reconstructed from different tracking methods. The visualization shows magnified patches of the green box.

| Error/Method | ICP | f2f | f2m | ours |
|---|---|---|---|---|
| RMSE | 0.1361 | 0.1177 | 0.1184 | **0.0902** |
| 1-NCC(3x3) | 0.7220 | 0.6124 | 0.6521 | **0.4692** |
| 1-NCC(5x5) | 0.6510 | 0.5430 | 0.5689 | **0.3877** |
| 1-NCC(7x7) | 0.5999 | 0.5043 | 0.5138 | **0.3491** |

Table 2: Photometric Errors of Tracking Methods

## 6. Experiments

We conducted qualitative and quantitative experiments to assess our method. All experiments use a calibrated Primesense Carmine RGBD sensor capturing 640x480 video at 30Hz for RGB and IR channels, synchronized with the same size depth video. Exposure time was set to be constant within each sequence. The system runs on a laptop with a NVIDIA GTX 980M GPU and an Intel i7 CPU.

### 6.1. HRT Tracking Evaluation

We evaluate the performance of our texture fusion method by measuring photometric error between the rendered and ground truth images for a held-out sample of frames. For comparison, we replace our HRT tracking with existing real-time camera tracking methods and evaluate the photometric error on a video of a highly textured Lambertian scene. We test RGBD tracking methods using a combination of frame-to-model Iterative Closest Point (ICP) [30], frame-to-frame (f2f) RGBD tracking [19, 34], and frame-to-model RGBD tracking (f2m) [40]. We did not implement loop closure techniques introduced in the state-of-the-art approaches such as that of [9, 40] as our target scenes are of smaller scale where tracking drift is limited.

Figure 8 compares the quality of texture fused using camera poses provided by each algorithm. Our high resolution frame-to-model tracking yields the sharpest result.

For a quantitative measure, we adopt the evaluation method introduced by [37] and compute the pixel-wise root-mean-square-error (RMSE) and 1-NCC error (for different patch sizes) between the ground-truth and rendered texture reconstruction. We refer to [37] for the precise procedure. Results can be found in Table 2.

### 6.2. Surface Light Field Rendering Results

To render a reconstructed model, we use $D$ represented as a texture map and implement a custom shader to evaluate $S$ with a HDR environment lighting which is captured from a consumer-grade 360 degree camera by taking several photos with varying exposure times [10]. Figure 9 compares the reconstructed surface light field of test scenes with real photographs. We highly encourage readers to watch the *supplementary video* for in-depth results.

These results demonstrate that our method can successfully recover surface light fields under a wide range of geometry, material, and lighting variations. The *Rooster* scene features challenging high-frequency textures and shiny ceramic surfaces, captured under office florescent lights. Our method can accurately reproduce sharp diffuse textures and specular highlights. The *Corncho* scene contains a bumpy specular vinyl surface captured near a large window. Notice the faithful soft-shadows and strong interreflections on the white floor expressed in the renderings. The *Dog* scene has significant self-occlusions. A bright directional lamp illuminates the object from a short distance, inducing strong specularities and sharp shadows. Our system faithfully reconstructs these effects. The *Bottle* scene contains a highly anisotropic brushed metal cylinder and a glossy black plastic cap. The results show that our system successfully recovers the vertically elongated specular highlights on the metal surface. Also notice that the relative specular reflectance of the two glossy materials are estimated correctly.

## 7. Limitations and Conclusions

Our system is unable to model near perfect mirrors or surfaces with micro-structure, due to inaccurate sensor shape measurement; e.g., tiny bumps on the *Corncho* model were not captured, causing noticeable difference in the sharpness of the reflection. Moreover, our surface light field formulation does not model colored metals such as bronze and omits specular interreflections.

We presented the first end-to-end system for computing surface light field object models using a hand-held, commodity RGBD sensor. We leverage a novel factorization of the surface light field that simplifies capture requirements, yet enables high quality results for a wide range of materials, including anisotropic ones. Our approach captures global illumination effects (like shadows and interreflections) and high resolution textures that greatly improve realism relative to prior interactive RGBD scanning methods.

## Acknowledgement

This work was supported by funding from the National Science Foundation grant IIS-1538618, Intel, Google, and the UW Reality Lab.

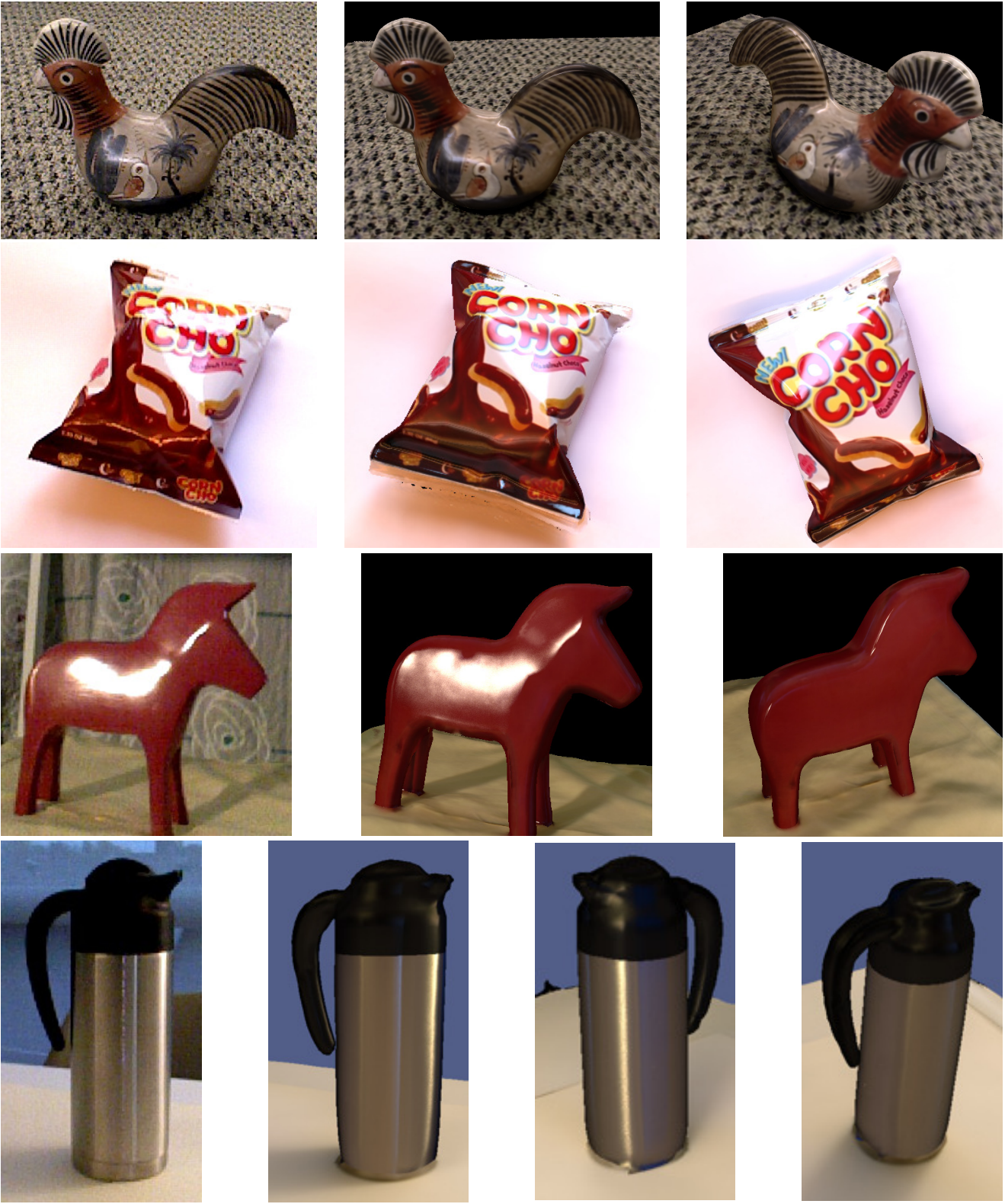

Figure 9: Ground truth images and renderings. First column: Ground truth photographs. Second column: Synthetic rendering of the same pose. Third and fourth column: Rendering of different camera poses.

# Appendix

### Details on Pairwise Term (Sec. 4.2)

The pairwise term incorporates both a photometric and a geometric smoothness prior:

$$\psi_{m,n}(y_m, y_n) = \mathbf{1}_{[y_m \neq y_n]} \left( \lambda_p \exp(-\theta_p ||\boldsymbol{c}_m - \boldsymbol{c}_n||^2) + \lambda_g \frac{(\boldsymbol{g}(m,n) + \epsilon)}{||\mathcal{N}(m) - \mathcal{N}(n)||^2} \right),$$

where $\lambda_p$, $\lambda_g$, $\theta_p$, $\epsilon$ are balancing parameters, and $\mathcal{N}$ and $\boldsymbol{c}$ are the vertex normal and color, respectively. Here, $\boldsymbol{g}(m,n)$ is an indicator function designed to penalize concave surface transitions as in [21]:

$$\boldsymbol{g}(m,n) = (\mathcal{N}(m) - \mathcal{N}(n)) \cdot (\mathcal{V}(m) - \mathcal{V}(n)) > 0,$$

where $\mathcal{V}$ is the vertex location. The optimal label $\boldsymbol{y}$ for $\Phi(\boldsymbol{y})$ is estimated via graph cuts [3].

### Details on Photometric Pose Refinement (Sec. 5.5)

On top of applying the high-pass filter to input and reference images, we mitigate the remaining specular effects by ignoring pixels with the gradient error of Equation (10) larger than 0.2. We preprocess the images with a Sobel filter to compute the gradient. We initialize $T^\circ$ with ICP and refine the pose by minimizing Equation (10) using a coarse-to-fine Gauss-Newton optimization operating on a three-level image pyramid.

### Additional Scene Reconstruction Result

We show a reconstruction result of another test scene (Figure 10). Note the high quality reconstruction of the multi-object scene with texture and specular highlights faithfully reproduced.

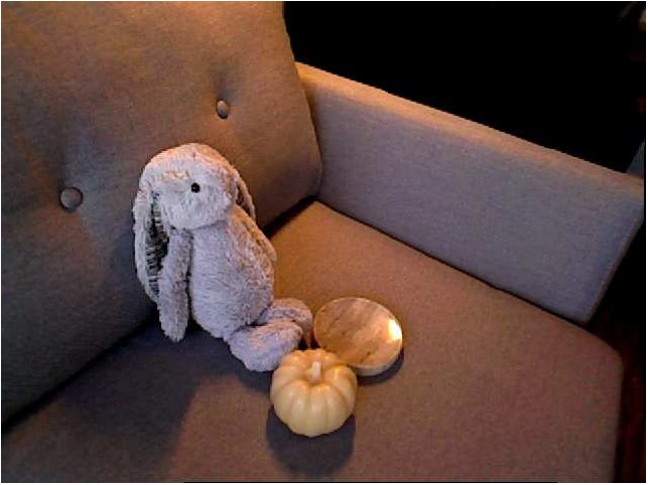

(a) Ground-truth

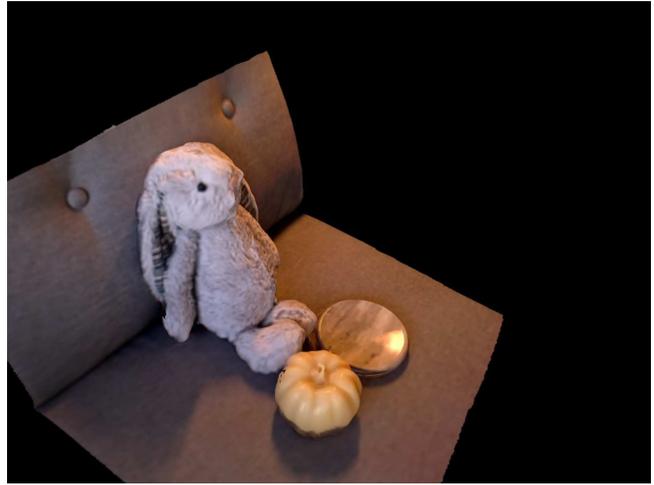

(b) Our Reconstruction

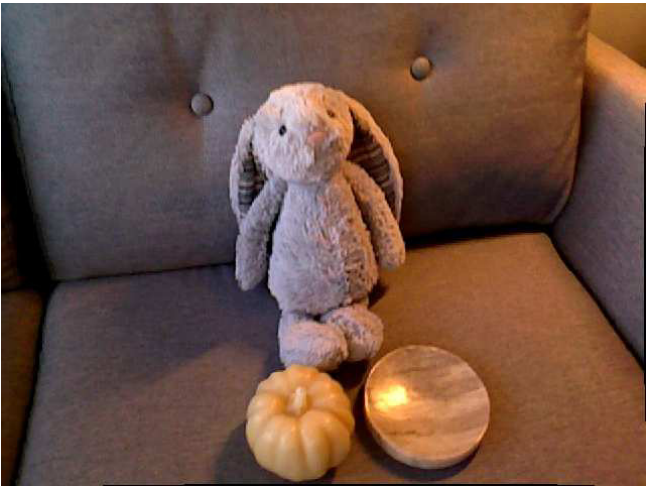

(c) Ground-truth

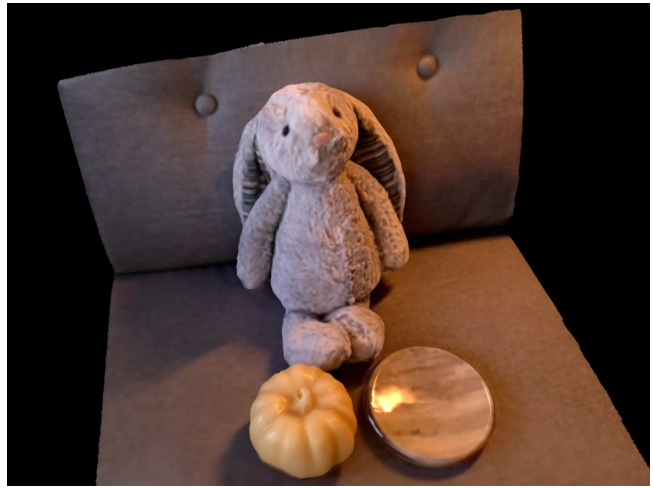

(d) Our Reconstruction

Figure 10: Ground-truth images and our surface light field renderings of the same camera pose.